\documentclass[11pt,letterpaper,implicit=false]{article} 
\usepackage{ijcnlp2017}
\usepackage{times}
\usepackage{latexsym}
\usepackage{graphicx}
\usepackage{amsmath}
\usepackage{amsfonts}
\usepackage{bm}
\usepackage{microtype}

\ijcnlpfinalcopy

\title{Input-to-Output Gate to Improve RNN Language Models}

\author{Sho Takase \hspace{1em} Jun Suzuki \hspace{1em} Masaaki Nagata \\
  NTT Communication Science Laboratories \\
  {\tt \{takase.sho, suzuki.jun, nagata.masaaki\}@lab.ntt.co.jp}}

\date{}

\begin{document}

\maketitle

\begin{abstract}
This paper proposes a reinforcing method that refines the output layers of existing Recurrent Neural Network (RNN) language models.
We refer to our proposed method as Input-to-Output Gate (IOG)\footnote{Our implementation is publicly available at \href{https://github.com/nttcslab-nlp/iog}{https://github.com/nttcslab-nlp/iog}.}.
IOG has an extremely simple structure, and thus, can be easily combined with any RNN language models.
Our experiments on the Penn Treebank and WikiText-2 datasets demonstrate that IOG consistently boosts the performance of several different types of current topline RNN language models.
\end{abstract}

\section{Introduction}
A neural language model is a central technology of recently developed neural architectures in the natural language processing (NLP) field.
For example, neural encoder-decoder models, which were successfully applied to various natural language generation tasks including machine translation~\cite{Sutskever:2014:SSL:2969033.2969173}, summarization~\cite{rush-chopra-weston:2015:EMNLP}, and dialogue~\cite{wen-EtAl:2015:EMNLP}, can be interpreted as conditional neural language models.
Moreover, word embedding methods, such as Skip-gram~\cite{NIPS2013_5021} and vLBL~\cite{NIPS2013_5165}, are also originated from neural language models that aim to handle much larger vocabulary and data sizes.
Thus, language modeling is a good benchmark task for investigating the general frameworks of neural methods in the NLP field.

In this paper, we address improving the performance on the language modeling task.
In particular, we focus on boosting the quality of existing Recurrent Neural Network (RNN) language models.
We propose the Input-to-Output Gate (IOG) method, which incorporates an additional gate function in the output layer of the selected RNN language model to refine the output.
One notable characteristic of IOG is that it can be easily incorporated in any RNN language models since it is designed to be a simple structure.
Our experiments on the Penn Treebank and WikiText-2 datasets demonstrate that IOG consistently boosts the performance of several different types of current topline RNN language models.
In addition, IOG achieves comparable scores to the state-of-the-art on the Penn Treebank dataset and outperforms the WikiText-2 dataset.

\section{RNN Language Model}
This section briefly overviews the RNN language models.
Hereafter, we denote a word sequence with length $T$, namely, $w_1, ..., w_T$ as $w_{1:T}$ for short.
Formally, a typical RNN language model computes the joint probability of word sequence $w_{1:T}$ by the product of the conditional probabilities of each timestep $t$:
\begin{align}
  p(w_{1:T}) = p(w_1)\prod_{t=1}^{T-1} p(w_{t+1} | w_{1:t}). \label{eq:probRNNLM}
\end{align}
$p(w_1)$ is generally assumed to be $1$ in this literature, that is, $p(w_1)\!=\!1$, and thus, we can ignore the calculation of this term (See the implementation of \newcite{DBLP:journals/corr/ZarembaSV14}\footnote{\href{https://github.com/wojzaremba/lstm}{{https://github.com/wojzaremba/lstm}}}, for example).
To estimate the conditional probability $p(w_{t+1} | w_{1:t})$, we apply RNNs.
Let $V$ be the vocabulary size, and let $P_{t} \in \mathbb{R}^{V}$ be the probability distribution of the vocabulary at timestep $t$.
Moreover, let $D_h$ and $D_e$ respectively be the dimensions of the hidden state and embedding vectors.
Then, the RNN language models predict $P_{t+1}$ by the following equation:
\begin{align}
  P_{t+1} &= {\rm softmax}(s_t), \label{eq:softmax} \\
  s_t &= W h_{t} + b, \label{eq:outlayer} \\
  h_t &= f(e_t, h_{t-1}), \label{eq:rnn} \\
  e_t &= E x_t, \label{eq:embed}
\end{align}
where $W \in \mathbb{R}^{V \times D_h}$ is a matrix, $b \in \mathbb{R}^{V}$ is a bias term, $E \in \mathbb{R}^{D_e \times V}$ is a word embedding matrix, $x_t \in \{0,1\}^{V}$ is a one-hot vector representing the word at timestep $t$, and $h_{t-1}$ is the hidden state at previous timestep $t-1$.
$h_t$ at timestep $t=0$ is defined as a zero vector, that is, $h_0 = \bm{0}$.
Let $f(\cdot)$ represent an abstract function of an RNN, which might be the Elman network~\cite{elman1990finding}, the Long Short-Term Memory (LSTM)~\cite{Hochreiter:1997:LSM:1246443.1246450}, the Recurrent Highway Network (RHN)~\cite{zilly2016recurrent}, or any other RNN variants.

\section{Input-to-Output Gate}
\begin{figure}[!t]
  \centering
  \includegraphics[width=6cm]{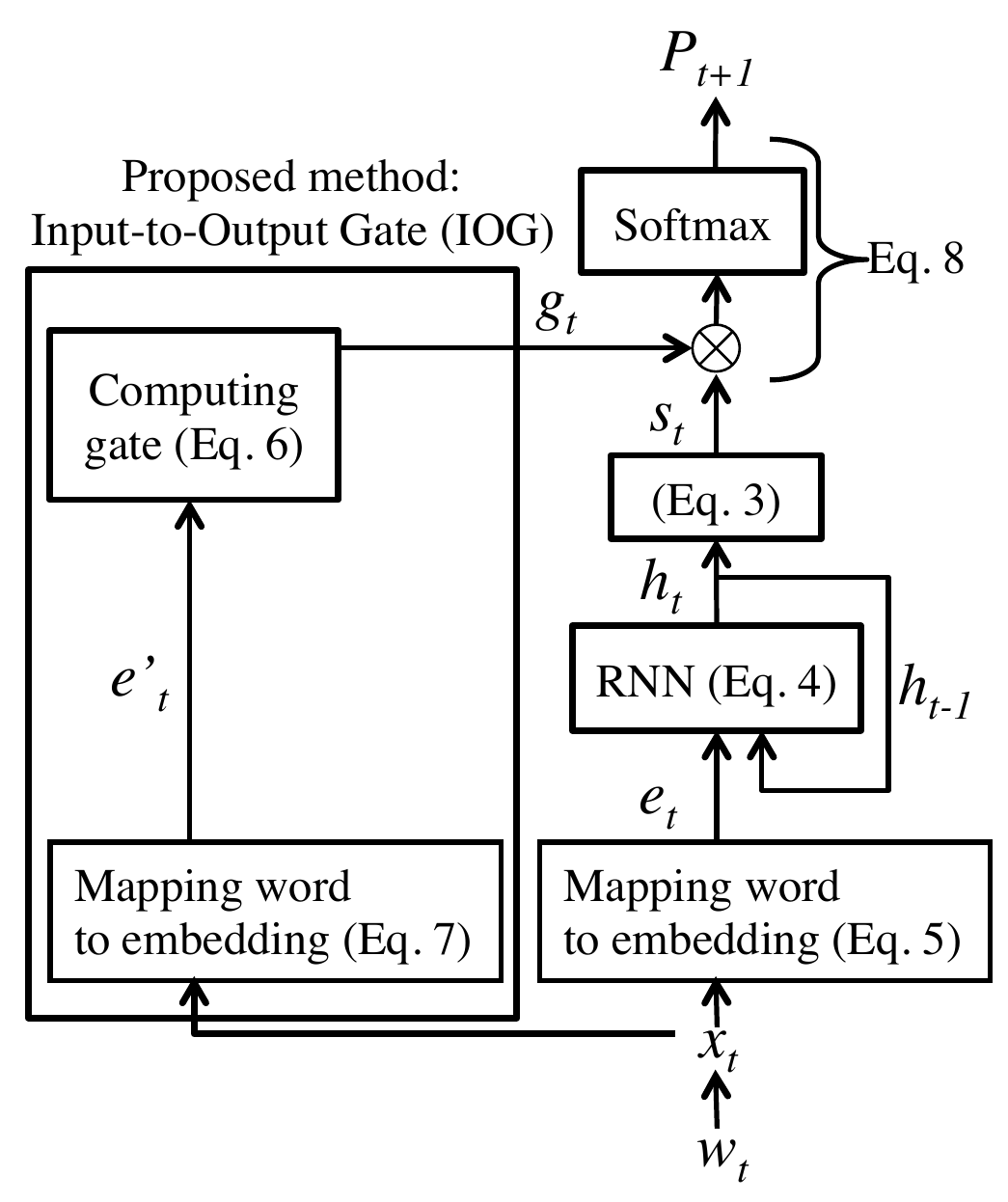}
   \caption{Overview of computing probability distribution.}
   \label{proposedMethod}
\end{figure}
In this section, we describe our proposed method: Input-to-Output Gate (IOG).
As illustrated in Figure~\ref{proposedMethod}, IOG adjusts the output of an RNN language model by the gate mechanism before computing the probability of the next word.
We expect that IOG will boost the probability of the word that may occur.
For example, a word followed by a preposition such as `of' is probably a noun.
Therefore, if the word at timestep $t$ is a preposition, IOG refines the output of a language model to raise the probabilities of nouns.

Formally, let $x_t$ be a one-hot vector representing $w_t$, IOG calculates the gate $g_t$ by the following equations:
\begin{align}
  g_t &= \sigma(W_g e'_t + b_g), \label{eq:proposedGate} \\
  e'_t &= E_g x_t.
\end{align}
Here, $W_g \in \mathbb{R}^{V \times D_g}$ is a matrix, $b_g \in \mathbb{R}^{V}$ is a bias term, and $E_g \in \mathbb{R}^{D_g \times V}$ is a word embedding matrix\footnote{We prepared different embeddings from those used in an RNN language model.}.
Then, we compute the probability distribution of the RNN language model by applying the above gate to the Equation (\ref{eq:softmax}) as follows:
\begin{align}
  P_{t+1} = {\rm softmax}(g_t \odot s_t), \label{eq:probWithIOG}
\end{align}
where $\odot$ represents the element-wise multiplication of two vectors.

\section{Experiments}

\begin{table}[!t]
  \centering
  \small
  \begin{tabular}{| l | c |} \hline
  Hyper-parameter & Selected value \\ \hline
  Embedding dimension $D_g$ & 300 \\
  Dropout rate & 50\% \\
  Optimization method & Adam \\
  Initial learning rate & 0.001 \\
  Learning rate decay & $1 / \sqrt{{\rm Epoch}} $ \\
  Max epoch & 5 \\ \hline
  \end{tabular}
  \caption{Hyper-parameters in training IOG.\label{tb:hyperparam}}
\end{table}

\subsection{Dataset}
We conducted word-level prediction experiments on the Penn Treebank (PTB)~\cite{Marcus:1993:BLA:972470.972475} and WikiText-2~\cite{DBLP:journals/corr/MerityXBS16} datasets.
The PTB dataset consists of 929k training words, 73k validation words, and 82k test words.
The WikiText-2 dataset consists of 2,088k training words, 217k validation words, and 245k test words.
\newcite{DBLP:conf/interspeech/MikolovKBCK10} and \newcite{DBLP:journals/corr/MerityXBS16} respectively published pre-processed PTB\footnote{\href{http://www.fit.vutbr.cz/~imikolov/rnnlm/}{{http://www.fit.vutbr.cz/~imikolov/rnnlm/}}} and WikiText-2\footnote{\href{https://einstein.ai/research/the-wikitext-long-term-dependency-language-modeling-dataset}{{https://einstein.ai/research/the-wikitext-long-term-dependency-language-modeling-dataset}}} datasets.
We used these pre-processed datasets for fair comparisons with previous studies.

\begin{table*}[!p]
  \centering
  \small
  \begin{tabular}{| l | c c c |} \hline
  Model & Parameters & Validation & Test \\ \hline
  LSTM (medium) \cite{DBLP:journals/corr/ZarembaSV14} $\dagger$ & 20M & 86.2 & 82.7 \\
  LSTM (medium, replication of \newcite{DBLP:journals/corr/ZarembaSV14}) & 20M & 87.1 & 84.0 \\
  + IOG (proposed) & 26M & 84.1 & 81.1 \\ \hline
  LSTM (large) \cite{DBLP:journals/corr/ZarembaSV14} $\dagger$ & 66M & 82.2 & 78.4 \\
  LSTM (large, replication of \newcite{DBLP:journals/corr/ZarembaSV14}) & 66M & 82.7 & 78.6 \\
  + IOG (proposed) & 72M & 78.5 & 75.5 \\ \hline
  Variational LSTM (medium) \cite{Gal2016Theoretically} $\dagger$ & 20M & 81.9 $\pm$ 0.2 & 79.7 $\pm$ 0.1 \\
  Variational LSTM (medium, replication of \newcite{Gal2016Theoretically}) & 20M & 82.8 & 79.1 \\
  + IOG (proposed) & 26M & 81.2 & 78.1 \\ \hline
  Variational LSTM (large) \cite{Gal2016Theoretically} $\dagger$ & 66M & 77.9 $\pm$ 0.3 & 75.2 $\pm$ 0.2 \\
  Variational LSTM (large, replication of \newcite{Gal2016Theoretically}) & 66M & 78.1 & 74.6 \\
  + IOG (proposed) & 72M & 76.9 & 74.1 \\ \hline
  Variational RHN (depth 8) \cite{zilly2016recurrent} $\dagger$ & 32M & 71.2 & 68.5 \\
  Variational RHN (depth 8, replication of \newcite{zilly2016recurrent}) & 32M & 72.1 & 68.9 \\
  + IOG (proposed) & 38M & 69.2 & 66.5 \\ \hline
  Variational RHN (depth 8, replication of \newcite{zilly2016recurrent}) + WT & 23M & 69.2 & 66.3 \\
  + IOG (proposed) & 29M & 67.0 & 64.4 \\ \hline
  Ensemble of 5 Variational RHNs & 160M & 66.1 & 63.1 \\
  + IOG (proposed) & 166M & 64.7 & 62.0 \\
  Ensemble of 10 Variational RHNs & 320M & 65.2 & 62.3 \\
  + IOG (proposed) & 326M & 64.1 & 61.4 \\ \hline \hline
  Neural cache model \cite{DBLP:journals/corr/GraveJU16} $\dagger$ & 21M & - & 72.1 \\
  Pointer Sentinel LSTM (medium) \cite{DBLP:journals/corr/MerityXBS16} $\dagger$ & 21M & 72.4 & 70.9 \\
  Variational LSTM (large) + WT + AL \cite{DBLP:journals/corr/InanKS16} $\dagger$ & 51M & 71.1 & 68.5 \\
  Variational RHN (depth 10) + WT \cite{press-wolf:2017:EACLshort} $\dagger$ & 24M & 68.1 & 66.0 \\
  Neural Architecture Search with base 8 \cite{45826} $\dagger$ & 32M & - & 67.9 \\
  Neural Architecture Search with base 8 + WT\cite{45826} $\dagger$ & 25M & - & 64.0 \\
  Neural Architecture Search with base 8 + WT \cite{45826} $\dagger$ & 54M & - & 62.4 \\ \hline \hline
  AWD LSTM + WT \cite{merityRegOpt} $\dagger$ & 24M & 60.0 & 57.3 \\
  AWD LSTM + WT (result by code of \newcite{merityRegOpt}\footnotemark) & 24M & 58.6 & 56.7 \\
  + IOG (proposed) & 30M & 58.5 & 56.7 \\ \hline
  AWD LSTM + WT + cache (size = 2000) \cite{merityRegOpt} $\dagger$ & 24M & 53.9 & {\bf 52.8} \\
  AWD LSTM + WT + cache (size = 500) & 24M & 53.4 & 53.0 \\
  + IOG (proposed) & 30M & {\bf 53.3} & 53.0 \\ \hline
  \end{tabular}
  \caption{Comparison between baseline models and the proposed method (represented as ``+ IOG'') on the Penn Treebank (PTB) dataset. $\dagger$ denotes results published in previous studies. The method with WT shared word embeddings ($E$ in the Equation (\ref{eq:embed})) with the weight matrix of the final layer ($W$ in the Equation (\ref{eq:outlayer})). AL denotes that the method used a previously proposed augmented loss function~\cite{DBLP:journals/corr/InanKS16}. \label{tb:perplexity}}

  \hspace{10truemm}

  \begin{tabular}{| l | c c c |} \hline
  Model & Parameters & Validation & Test \\ \hline
  LSTM (medium, replication of \newcite{DBLP:journals/corr/ZarembaSV14}) & 50M & 102.2 & 96.2 \\
  + IOG (proposed) & 70M & 99.2 & 93.8 \\ \hline
  Variational LSTM (medium, replication of \newcite{Gal2016Theoretically}) & 50M & 97.2 & 91.8 \\
  + IOG (proposed) & 70M & 95.9 & 91.0 \\ \hline
  Variational LSTM (medium) + cache (size = 2000) & 50M & 69.6 & 66.1 \\
  + IOG (proposed) & 70M & 69.3 & 65.9 \\ \hline \hline
  Pointer Sentinel LSTM \cite{DBLP:journals/corr/MerityXBS16} $\dagger$ & 51M \footnotemark & 84.8 & 80.8 \\
  Neural cache model (size = 100) \cite{DBLP:journals/corr/GraveJU16} $\dagger$ & 42M & - & 81.6 \\
  Neural cache model (size = 2000) \cite{DBLP:journals/corr/GraveJU16} $\dagger$ & 42M & - & 68.9 \\ \hline \hline
  AWD LSTM + WT \cite{merityRegOpt} $\dagger$ & 33M & 68.6 & 65.8 \\
  AWD LSTM + WT (result by code of \newcite{merityRegOpt}) & 33M & 68.6 & 65.8 \\
  + IOG (proposed) & 53M & 68.6 & 65.9 \\ \hline
  AWD LSTM + WT + cache (size = 3785) \cite{merityRegOpt} $\dagger$ & 33M & 53.8 & 52.0 \\
  AWD LSTM + WT + cache (size = 3785) & 33M & {\bf 53.5} & {\bf 51.7} \\
  + IOG (proposed) & 53M & 53.6 & {\bf 51.7} \\ \hline
  \end{tabular}
  \caption{Comparison between baseline models and the proposed method (represented as ``+ IOG'') on the WikiText-2 dataset. $\dagger$ denotes results published in previous studies.\label{tb:perplexityOnWikitext}}
\end{table*}

\subsection{Training Procedure}
\label{trainingSetting}
For the PTB dataset, we prepared a total of 5 RNN language models as our baseline models.
First, we replicated LSTM with dropout and LSTM with variational inference based dropout, which we refer to as ``LSTM'' and ``Variational LSTM'', respectively.
Following \newcite{DBLP:journals/corr/ZarembaSV14} and \newcite{Gal2016Theoretically}, we prepared the medium setting (2-layer LSTM with 650 dimensions for each layer), and the large setting (2-layer LSTM with 1500 dimensions for each layer) for each LSTM.
We also replicated ``Variational RHN'' with a depth of 8 described in \newcite{zilly2016recurrent}.
For the WikiText-2 dataset, we prepared the medium setting standard and variational LSTMs as our baselines, which are identical as those used in~\newcite{DBLP:journals/corr/MerityXBS16}.

After reproducing the baselines, we incorporated IOG with those models.
Table~\ref{tb:hyperparam} summarizes the hyper-parameters used for training the IOG.
During training IOG, we fixed the parameters of the RNN language models to avoid over-fitting.

\subsection{Results}

\footnotetext[6]{In contrast to other comparisons, we used the following implementation by the authors: \href{https://github.com/salesforce/awd-lstm-lm}{https://github.com/salesforce/awd-lstm-lm}}
\footnotetext[7]{The number of parameters is different from the one described in \newcite{DBLP:journals/corr/MerityXBS16}. We guess that they do not consider the increase of the vocabulary size.}

We show the perplexities of the baselines and those combined with IOG for the PTB in Table~\ref{tb:perplexity}, and for the WikiText-2 in Table~\ref{tb:perplexityOnWikitext}.
These tables, which contain both the scores reported in the previous studies and those obtained by our reproduced models, indicate that IOG reduced the perplexity.
In other words, IOG boosted the performance of the baseline models.
We emphasize that IOG is not restricted to a neural architecture of a language model because it improved the RHN and LSTM performances.

In addition to the comparison with the baselines, Table~\ref{tb:perplexity} and Table~\ref{tb:perplexityOnWikitext} contain the scores published in previous studies.
\newcite{DBLP:journals/corr/MerityXBS16} and \newcite{DBLP:journals/corr/GraveJU16} proposed similar methods.
Their methods, which are called ``cache mechanism'' (or `pointer'), keep multiple hidden states at past timesteps to select words from previous sequences.
\newcite{DBLP:journals/corr/InanKS16} and \newcite{press-wolf:2017:EACLshort} introduced a technique that shares word embeddings with the weight matrix of the final layer (represented as `WT' in Table~\ref{tb:perplexity}).
\newcite{DBLP:journals/corr/InanKS16} also proposed using word embeddings to augment loss function (represented as `AL' in Table~\ref{tb:perplexity}).
\newcite{45826} adopted RNNs and reinforcement learning to automatically construct a novel RNN architecture.
We expect that IOG will improve these models since it can be combined with any RNN language models.
In fact, Table~\ref{tb:perplexity} and Table~\ref{tb:perplexityOnWikitext} demonstrate that IOG enhanced the performance even when the RNN language model was combined with `WT' or the cache mechanism.

Table~\ref{tb:perplexity} also shows the scores in the ensemble settings.
Model ensemble techniques are widely used for further improving the performance of neural networks.
In this experiment, we employed a simple ensemble technique: using the average of the output probability distributions from each model as output.
We computed the probability distribution $P_{t+1}$ on the ensemble of the $M$ models as follows:
\begin{align}
  P_{t+1} = \frac{1}{M} \sum_{m=1}^{M} {}_{m}P_{t+1},
\end{align}
where ${}_{m}P_{t+1}$ represents the probability distribution predicted by the $m$-th model.
In the ensemble setting, we applied only one IOG to the multiple models.
In other words, we used the same IOG for computing the probability distributions of each language model, namely, computing the Equation (\ref{eq:probWithIOG}).
Table~\ref{tb:perplexity} describes that 5 and 10 model ensemble of Variational RHNs outperformed the single model by more than 5 in perplexity.
Table~\ref{tb:perplexity} shows that IOG reduced the perplexity of the ensemble models.
Remarkably, even though the 10 Variational RHN ensemble achieved the state-of-the-art performance on the PTB dataset, IOG improved the performance by about 1 in perplexity\footnote{This result was the state-of-the-art score at the submission deadline of IJCNLP 2017, i.e., July 7, 2017, but \newcite{merityRegOpt} surpassed it on Aug 7, 2017. We mention the effect of IOG on their method in the following paragraph.}.

In addition, as additional experiments, we incorporated IOG with the latest method, which was proposed after the submission deadline of IJCNLP 2017.
\newcite{merityRegOpt} introduced various regularization and optimization techniques such as DropConnect~\cite{wan2013regularization} and averaged stochastic gradient descent~\cite{polyak1992acceleration} to the LSTM language model.
They called their approach AWD LSTM, which is an abbreviation of averaged stochastic gradient descent weight-dropped LSTM.
Table~\ref{tb:perplexity} and Table~\ref{tb:perplexityOnWikitext} indicate the results on the PTB and the WikiText-2 respectively.
These tables show that IOG was not effective to AWD LSTM.
Perhaps, the reason is that the perplexity of AWD LSTM is close to the best performance of the simple LSTM architecture.
We also note that IOG did not have any harmful effect on the language models because it maintained the performances of AWD LSTM with `WT' and the cache mechanism.
Moreover, incorporating IOG is much easier than exploring the best regularization and optimization methods for each RNN language model.
Therefore, to improve the performance, we recommend combining IOG before searching for the best practice.

\subsection{Discussion}
\begin{table}[!t]
  \centering
  \small
  \begin{tabular}{| l | c c |} \hline
  Model & Diff & Test \\ \hline
  Variational RHN (replicate) & - & 68.9 \\
  Variational RHN + IOG (proposed) & - & {\bf 66.5} \\
  Variational RHN + IOG with hidden & +0.8M & 75.6 \\
  Variational RHN + LSTM gate & +0.7M & 68.1 \\ \hline
  \end{tabular}
  \caption{Comparison among architectures for computing the output gate on the PTB dataset. The column `Diff' shows increase of parameters from IOG (proposed).\label{tb:archtecture}}
\end{table}

Although IOG consists only of word embeddings and one weight matrix, the experimental results were surprisingly good.
One might think that more sophisticated architectures can provide further improvements.
To investigate this question, we examined two additional architectures to compute the output gate $g_t$ in the Equation (\ref{eq:proposedGate}).

The first one substituted the calculation of the gate function $g_t$ by the following $g'_t$:
\begin{align}
 g'_t &= \sigma(W'_g [h_t, e'_t] + b_g),
\end{align}
where $W'_g\in\mathbb{R}^{V \times (D_h + D_g)}$, and $[h_t, e'_t]$ represents the concatenation of the hidden state $h_t$ of RHN and embeddings $e'_t$ used in IOG.
We refer to this architecture as ``+ IOG with hidden''.

The second one similarly substituted $g_t$ by the following $g''_t$:
\begin{align}
  g''_t &= \sigma(W_g h'_t + b_g), \\
  h'_t &= f'(e'_t, h'_{t-1}), 
\end{align}
where $f'(\cdot)$ is the 1-layer LSTM in our experiments.
We set the dimension of the LSTM hidden state to 300, that is, $D_g\!=\!300$, and the other hyper-parameters remained as described in Section~\ref{trainingSetting}.
We refer to the second one as ``+ LSTM gate''.

Table~\ref{tb:archtecture} shows the results of the above two architectures on the PTB dataset.
IOG clearly outperformed the other more sophisticated architectures.
This fact suggests that (1) incorporating additional architectures does not always improve the performance, and (2) not always become better even if it is a sophisticated architecture.
We need to carefully design an architecture that can provide complementary (or orthogonal) information to the baseline RNNs.

\begin{table}[!t]
  \centering
  \small
  \begin{tabular}{| l | l |} \hline
  Input word & Top 5 weighted words \\ \hline
  of & security, columbia, steel, irs, thrift \\
  in & columbia, ford, order, labor, east \\
  go & after, through, back, on, ahead \\
  attention & was, than, $\langle$eos$\rangle$, from, to \\
  whether & to, she, estimates, i, ual \\ \hline
  \end{tabular}
  \caption{Top 5 weighted words for each input word on the PTB experiment.}
  \label{tb:weightedWords}
\end{table}

In addition, to investigate the mechanism of IOG, we selected particular words, and listed the top 5 weighted words given each selected word as input in Table~\ref{tb:weightedWords}\footnote{In this exploration, we excluded words occurring fewer than 100 times in the corpus to remove noise.}.
IOG gave high weights to nouns when the input word was a preposition: `of' and `in'.
Moreover, IOG encouraged outputting phrasal verbs such as ``go after''.
These observations generally match human intuition.

\section{Conclusion}
We proposed Input-to-Output Gate (IOG), which refines the output of an RNN language model by the gate mechanism.
IOG can be incorporated in any RNN language models due to its simple structure.
In fact, our experimental results demonstrated that IOG improved the performance of several different settings of RNN language models.
Furthermore, the experimental results indicate that IOG can be used with other techniques such as ensemble.


\begin{thebibliography}{}
\expandafter\ifx\csname natexlab\endcsname\relax\def\natexlab#1{#1}\fi

\bibitem[{Elman(1990)}]{elman1990finding}
Jeffrey~L Elman. 1990.
\newblock {Finding Structure in Time}.
\newblock {\em Cognitive science\/} 14(2):179--211.

\bibitem[{Gal and Ghahramani(2016)}]{Gal2016Theoretically}
Yarin Gal and Zoubin Ghahramani. 2016.
\newblock {A Theoretically Grounded Application of Dropout in Recurrent Neural
  Networks}.
\newblock In {\em Advances in Neural Information Processing Systems 29 (NIPS
  2016)\/}.

\bibitem[{Grave et~al.(2017)Grave, Joulin, and
  Usunier}]{DBLP:journals/corr/GraveJU16}
Edouard Grave, Armand Joulin, and Nicolas Usunier. 2017.
\newblock {Improving Neural Language Models with a Continuous Cache}.
\newblock In {\em 5th International Conference on Learning Representations
  (ICLR 2017)\/}.

\bibitem[{Hochreiter and
  Schmidhuber(1997)}]{Hochreiter:1997:LSM:1246443.1246450}
Sepp Hochreiter and J\"{u}rgen Schmidhuber. 1997.
\newblock {Long Short-Term Memory}.
\newblock {\em Neural Computation\/} 9(8):1735--1780.

\bibitem[{Inan et~al.(2016)Inan, Khosravi, and
  Socher}]{DBLP:journals/corr/InanKS16}
Hakan Inan, Khashayar Khosravi, and Richard Socher. 2016.
\newblock {Tying Word Vectors and Word Classifiers: {A} Loss Framework for
  Language Modeling}.
\newblock In {\em Proceedings of the 5th International Conference on Learning
  Representations (ICLR 2017)\/}.

\bibitem[{Marcus et~al.(1993)Marcus, Marcinkiewicz, and
  Santorini}]{Marcus:1993:BLA:972470.972475}
Mitchell~P. Marcus, Mary~Ann Marcinkiewicz, and Beatrice Santorini. 1993.
\newblock {Building a Large Annotated Corpus of English: The Penn Treebank}.
\newblock {\em Computational Linguistics\/} 19(2):313--330.

\bibitem[{Merity et~al.(2017{\natexlab{a}})Merity, Keskar, and
  Socher}]{merityRegOpt}
Stephen Merity, Nitish~Shirish Keskar, and Richard Socher. 2017{\natexlab{a}}.
\newblock {Regularizing and Optimizing LSTM Language Models}.
\newblock {\em arXiv preprint arXiv:1708.02182\/} .

\bibitem[{Merity et~al.(2017{\natexlab{b}})Merity, Xiong, Bradbury, and
  Socher}]{DBLP:journals/corr/MerityXBS16}
Stephen Merity, Caiming Xiong, James Bradbury, and Richard Socher.
  2017{\natexlab{b}}.
\newblock {Pointer Sentinel Mixture Models}.
\newblock In {\em Proceedings of the 5th International Conference on Learning
  Representations (ICLR 2017)\/}.

\bibitem[{Mikolov et~al.(2010)Mikolov, Karafi{\'{a}}t, Burget, Cernock{\'{y}},
  and Khudanpur}]{DBLP:conf/interspeech/MikolovKBCK10}
Tomas Mikolov, Martin Karafi{\'{a}}t, Luk{\'{a}}s Burget, Jan Cernock{\'{y}},
  and Sanjeev Khudanpur. 2010.
\newblock {Recurrent Neural Network based Language Model}.
\newblock In {\em Proceedings of the 11th Annual Conference of the
  International Speech Communication Association (INTERSPEECH 2010)\/}. pages
  1045--1048.

\bibitem[{Mikolov et~al.(2013)Mikolov, Sutskever, Chen, Corrado, and
  Dean}]{NIPS2013_5021}
Tomas Mikolov, Ilya Sutskever, Kai Chen, Greg~S Corrado, and Jeff Dean. 2013.
\newblock {Distributed Representations of Words and Phrases and their
  Compositionality}.
\newblock In {\em Advances in Neural Information Processing Systems 26 (NIPS
  2013)\/}, pages 3111--3119.

\bibitem[{Mnih and Kavukcuoglu(2013)}]{NIPS2013_5165}
Andriy Mnih and Koray Kavukcuoglu. 2013.
\newblock {Learning Word Embeddings Efficiently with Noise-Contrastive
  Estimation}.
\newblock In {\em Advances in Neural Information Processing Systems 26 (NIPS
  2013)\/}, pages 2265--2273.

\bibitem[{Polyak and Juditsky(1992)}]{polyak1992acceleration}
Boris~T Polyak and Anatoli~B Juditsky. 1992.
\newblock {Acceleration of Stochastic Approximation by Averaging}.
\newblock {\em SIAM Journal on Control and Optimization\/} 30(4):838--855.

\bibitem[{Press and Wolf(2017)}]{press-wolf:2017:EACLshort}
Ofir Press and Lior Wolf. 2017.
\newblock {Using the Output Embedding to Improve Language Models}.
\newblock In {\em Proceedings of the 15th Conference of the European Chapter of
  the Association for Computational Linguistics (EACL 2017)\/}. pages 157--163.

\bibitem[{Rush et~al.(2015)Rush, Chopra, and
  Weston}]{rush-chopra-weston:2015:EMNLP}
Alexander~M. Rush, Sumit Chopra, and Jason Weston. 2015.
\newblock {A Neural Attention Model for Abstractive Sentence Summarization}.
\newblock In {\em Proceedings of the 2015 Conference on Empirical Methods in
  Natural Language Processing (EMNLP 2015)\/}. pages 379--389.

\bibitem[{Sutskever et~al.(2014)Sutskever, Vinyals, and
  Le}]{Sutskever:2014:SSL:2969033.2969173}
Ilya Sutskever, Oriol Vinyals, and Quoc~V. Le. 2014.
\newblock {Sequence to Sequence Learning with Neural Networks}.
\newblock In {\em Advances in Neural Information Processing Systems 27 (NIPS
  2014)\/}. pages 3104--3112.

\bibitem[{Wan et~al.(2013)Wan, Zeiler, Zhang, Cun, and
  Fergus}]{wan2013regularization}
Li~Wan, Matthew Zeiler, Sixin Zhang, Yann~L Cun, and Rob Fergus. 2013.
\newblock {Regularization of Neural Networks using DropConnect}.
\newblock In {\em Proceedings of the 30th International Conference on Machine
  Learning (ICML 2013)\/}. pages 1058--1066.

\bibitem[{Wen et~al.(2015)Wen, Gasic, Mrk\v{s}i\'{c}, Su, Vandyke, and
  Young}]{wen-EtAl:2015:EMNLP}
Tsung-Hsien Wen, Milica Gasic, Nikola Mrk\v{s}i\'{c}, Pei-Hao Su, David
  Vandyke, and Steve Young. 2015.
\newblock {Semantically Conditioned LSTM-based Natural Language Generation for
  Spoken Dialogue Systems}.
\newblock In {\em Proceedings of the 2015 Conference on Empirical Methods in
  Natural Language Processing (EMNLP 2015)\/}. pages 1711--1721.

\bibitem[{Zaremba et~al.(2014)Zaremba, Sutskever, and
  Vinyals}]{DBLP:journals/corr/ZarembaSV14}
Wojciech Zaremba, Ilya Sutskever, and Oriol Vinyals. 2014.
\newblock {Recurrent Neural Network Regularization}.
\newblock In {\em Proceedings of the 2nd International Conference on Learning
  Representations (ICLR 2014)\/}.

\bibitem[{Zilly et~al.(2017)Zilly, Srivastava, Koutn{\'\i}k, and
  Schmidhuber}]{zilly2016recurrent}
Julian~Georg Zilly, Rupesh~Kumar Srivastava, Jan Koutn{\'\i}k, and J{\"u}rgen
  Schmidhuber. 2017.
\newblock {Recurrent Highway Networks}.
\newblock {\em Proceedings of the 34th International Conference on Machine
  Learning (ICML 2017)\/} pages 4189--4198.

\bibitem[{Zoph and Le(2017)}]{45826}
Barret Zoph and Quoc~V. Le. 2017.
\newblock {Neural Architecture Search with Reinforcement Learning}.
\newblock In {\em Proceedings of the 5th International Conference on Learning
  Representations (ICLR 2017)\/}.

\end{thebibliography}
\bibliographystyle{ijcnlp2017}

\end{document}